\documentclass[conference]{IEEEtran}
\IEEEoverridecommandlockouts
\usepackage{cite}
\usepackage{amsmath,amssymb,amsfonts}
\usepackage{algorithmic}
\usepackage{graphicx}
\usepackage{textcomp}
\usepackage{xcolor}
\usepackage{multirow}

\usepackage[utf8]{inputenc} 
\usepackage[T1]{fontenc}    
\usepackage{hyperref}       
\usepackage{url}            
\usepackage{booktabs}       
\usepackage{amsfonts}       
\usepackage{amsmath}
\usepackage{nicefrac}       
\usepackage{microtype}      
\usepackage{optidef}
\usepackage{subcaption}
\usepackage{multirow}

\def\BibTeX{{\rm B\kern-.05em{\sc i\kern-.025em b}\kern-.08em
    T\kern-.1667em\lower.7ex\hbox{E}\kern-.125emX}}

\begin{document}

\title{Drug-Target Indication Prediction by Integrating End-to-End Learning and Fingerprints
\thanks{This work was supported by SipingSoft Co. Ltd., Chengdu, P. R. China}
}

\author{\IEEEauthorblockN{Brighter Agyemang*}
\IEEEauthorblockA{\textit{School of Computer Science and Software Engineering} \\
\textit{University of Electronic Science and Technology of China}\\
Chengdu, P. R. China \\
brighteragyemang@gmail.com}
\and
\IEEEauthorblockN{Wei-Ping Wu}
\IEEEauthorblockA{\textit{School of Computer Science and Software Engineering} \\
\textit{University of Electronic Science and Technology of China}\\
Chengdu, P. R. China \\
wei-ping.wu@uestc.edu.cn}
\and
\IEEEauthorblockN{Michael Y. Kpiebaareh}
\IEEEauthorblockA{\textit{School of Computer Science and Software Engineering} \\
\textit{University of Electronic Science and Technology of China}\\
Chengdu, P. R. China \\
kpiebaareh@yahoo.com}
\and
\IEEEauthorblockN{Ebenezer Nanor}
\IEEEauthorblockA{\textit{School of Computer Science and Software Engineering} \\
\textit{University of Electronic Science and Technology of China}\\
Chengdu, P. R. China \\
sireben21@gmail.com}
}

\maketitle

\begin{abstract}
Computer-Aided Drug Discovery research has proven to be a promising direction in drug discovery. In recent years, Deep Learning approaches have been applied to problems in the domain such as Drug-Target Interaction Prediction and have shown improvements over traditional screening methods. 
An existing challenge is how to represent compound-target pairs in deep learning models. While several representation methods exist, such descriptor schemes tend to complement one another in many instances, as reported in the literature. 
In this study, we propose a multi-view architecture trained adversarially to leverage this complementary behavior by integrating both differentiable and predefined molecular descriptors. We conduct experiments on clinically relevant benchmark datasets to demonstrate the potential of our approach.
\end{abstract}


\section{Introduction}
Over the years, drug discovery has predominantly transformed from crude and serendipitous characteristics to a well-structured and rational scientific paradigm. The paradox, however, is that the growth experienced in the domain has not made the identification of drugs any easier as attested to by the high attrition rates and huge budgets typically involved in the process\cite{DiMasi2003}. Traditionally, \emph{in vitro} screening experiments are conducted in order to identify new interactions. However, considering that there are about $10^{60}$ synthetically feasible compounds, \emph{in silico} or virtual screening alternatives are mostly used for this process \cite{Polishchuk2013}. 
The two main \emph{in silico} approaches to Drug-Target Interaction (DTI) prediction are docking simulations and machine learning approaches. Docking simulations utilize compound and target conformations to discover binding sites whereas machine learning methods are based on using features of compounds and/or targets, or their similarities. 
 
A central aspect in applying these computational models is the featurization of compounds and biological targets into numerical vectors. This is achieved using molecular descriptors which encode properties such as bonds, valence structures, sequence information, and other related properties \cite{Rifaioglu2018}. Digitally, 2D structures of compounds are represented using line notations such the Simplified Molecular Input Line Entry System (SMILES) \cite{Weininger1988} whereas targets are represented using sequences, and/or their conformations when available. These digital forms are then used by toolkits (e.g. RDKit \cite{Landrum2006}) to derive the molecular descriptors.

However, there exist several different kinds of molecular descriptors in the literature with Extended Connectivity Fingerprints (ECFP) being one of the widely used descriptors. Considering that for a given compound, different descriptors produce different properties which affect model performance, this makes the choice a featurization method in model development a significant one \cite{CERETOMASSAGUE201558,Kogej2006}. 
In certain cases, the performance of certain descriptors tend to be task related \cite{DUAN2010157,Rifaioglu2018}. To this end, it is usual in the domain to find different descriptors being combined as input to a model, an approach referred to as Joint Multi-Modal Learning~\cite{Baltrusaitis2019}. Nonetheless, the unimodal features that are used in constructing the joint representations tend to be constructed from pre-determined descriptors.

In this study, we propose an integrated view predictive architecture that is trained adversarially~\cite{Lotter2015} for predicting compound-target binding affinities. The motivation is that, these descriptors tend to complement one another in many cases and that the different modalities could provide an in-depth perspective about a sample~\cite{Rifaioglu2018}. Additionally, the enormity of the chemical space connotes that task-specific representation of compounds is a desideratum of DTI prediction research~\cite{Wu2018}. Therefore, our departure from constructing joint representations exclusively with predefined descriptors to leveraging differential feature learning and such predefined descriptors dovetails into the concept of tailoring DTI tasks to their input space.

The subsequent sections are organized as follows: section \ref{sec:relwrk} highlights related work of our study, section \ref{sec:model} discusses our proposed model, section \ref{sec:exp} presents the experiments design of our study. We also discuss the results of our experiments in section~\ref{sec:res_discu} and draw our conclusions in section~\ref{sec:conc}.

\section{Related Work}\label{sec:relwrk}

The concept of utilizing multiple unimodal descriptors toward predicting bioactivity has been well studied in the literature. Since different descriptors tend to represent different properties of compounds~\cite{CERETOMASSAGUE201558}, integrating these descriptors has been considered in several studies. In \cite{Sawada2014}, 18 different chemical descriptors are benchmarked on DTI prediction tasks and their findings reveal that integrating multiple descriptors typically improves model performance. The works of Soufan et al~\cite{Soufan2016,Soufan2015} also corroborate this observation.
Hence, the combination enables the participating descriptors to create informative feature vectors. 

While these studies espouse integrating multiple chemical descriptors, existing studies have mostly employed predefined feature sets, such as structure-based fingerprints and pharmacophore descriptors proposed by experts in the domain, on DTI prediction problems. Recent studies have also proposed end-to-end compound descriptor learning functions toward ensuring a closer relationship between the learning objective and the input space ~\cite{Wu2018, Gomes2017,Kearnes2016}. Although several studies have approached DTI prediction as a binary classification problem~\cite{Lee2019}, the nature of bioactivity is deemed to be continuous~\cite{Pahikkala2015a}. In \cite{Feng2018}, a DL model using ECFP (with diameter 4) is compared to a Molecular Graph Convolution (GraphConv) model on predicting binding affinities. In both cases, target information, in the form of Protein Sequence Composition (PSC), is combined with the descriptor of a compound for predicting the binding strengths.  The work in~\cite{Feng2018} also shows that using DL methods for predicting bioactivity generally leads to better performance than kernel- and gradient boosting machines-based methods~\cite{Pahikkala2015a,He2017}.

We propose a predictive generative adversarial network~\cite{Lotter2015} architecture,  for leveraging the complementary relationship between the seemingly disparate featurization methods in the domain. We show that our approach generally improves the findings in~\cite{Feng2018}.

\section{Model}\label{sec:model}
\subsection{Problem Formulation}
We consider the problem of predicting a real-valued binding affinity $y_i$ between a given compound $c_i$ and target $p_i$, $i\in\mathbb{R}$. The compound $c_i$ is represented as a SMILES~\cite{Weininger1988} string whereas the target $p_i$ is represented as an amino acid sequence. The target feature vector is constructed using Protein Sequence Composition (PSC), which comprises of the Amino Acid Composition (AAC), Dipeptide Composition (DC), and Tripeptide Composition (TC), using ProPy~\cite{Cao2013}. The SMILES string of $c_i$ is an encoding of a chemical graph structure $d_i=\{V_i,E_i\}$, where $V_i$ is the set of atoms constituting $c_i$ and $E_i$ is a set of undirected chemical bonds between these atoms. While predefined fingerprints are computed by directly examining $d_i$, descriptor learning functions like GraphConv~\cite{Altae-Tran2017} take as input $d_i$ and learn the numerical representation of $c_i$ using backpropagation.

\subsection{Proposed Architecture}
\begin{figure*}
    \centering
    \fbox{\includegraphics[width=.75\linewidth]{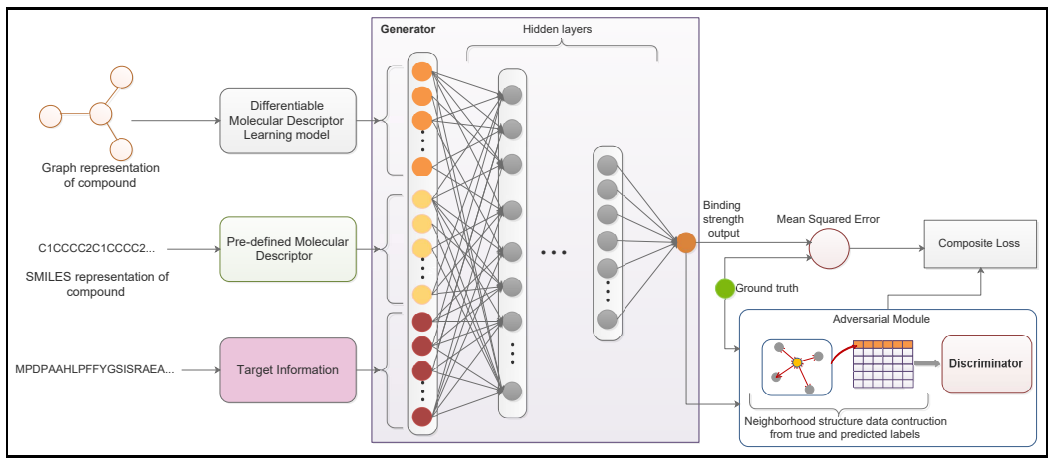}}
    \caption{An integrated architecture of different views for DTI prediction using a Predictive GAN approach.}
    \label{fig:integrated}
\end{figure*}

The proposed Integrated Views Predictive Generative Adverserial Network (IVPGAN) for DTI, is shown in figure~\ref{fig:integrated}. Given a predefined descriptor vector $v_i^c$ of $c_i$, chemical graph structure $d_i$, and the PSC vector $v_i^p$ of $p_i$, we optimize the following mean squared error (MSE) loss,

\begin{equation}
    \operatorname*{argmin}_{\theta^f,\theta^g} \sum_{i=0}^N (y_i - f([v_i^c,g(d_i;\theta^g),v_i^p];\theta^f))^2
    \label{eq:mse}
\end{equation}
where $\theta^f$ and $\theta^g$ are trainable parameters. 

Thus, we form a joint representation of the query entities $c_i$ and $p_i$ by concatenating the predefined molecular descriptor, the outputs of the parameterized descriptor learning function $g(\cdot;\theta^g)$, and the PSC feature vector of $p_i$ We refer to this joint representation as the Combined Input Vector (CIV). Since equation~\ref{eq:mse} is able to estimate the bias and variance of an estimator, this makes it a good fit for real-valued models.

However, the squared loss is sensitive to the overall departure of the samples and tends to represent the output distribution by placing masses in parts of the space with low densities. In computer vision problems, this results in a blurring effect. Therefore, we follow Lotter et. al~\cite{Lotter2015} to train our DTI prediction model adversarially by applying Adversarial Loss (AL). Specifically, we view the model accepting CIV input as \emph{generating} a binding strength between a given pair, as against a vanilla prediction model. This perspective enables us to leverage techniques in Generative Adverserial Networks (GANS)~\cite{Goodfellow14generativeadversarial} to mitigate the aforementioned problem of equation~\ref{eq:mse}. Also, models trained adverserially are able to learn the structured patterns in a distribution~\cite{Lotter2015}.

To this end, given a set of generated binding strengths $\mathbf{\hat{y}}=\{\hat{y}_1,\hat{y}_2,...,\hat{y}_N\}$ produced by a generator $G$ and their corresponding ground truths $\mathbf{y}=\{y_1,y_2,...,y_N\}$, we construct two neighborhood alignment datasets for the adversarial training phase.

Let $Y=\{\mathbf{\hat{y}},\mathbf{y}\}$ be the set of predicted and label vectors. We construct the data matrix $M^{(l)}$ from $Y^{(l)}$, $0<l\leq 2$, as:

\begin{equation}
    M^{(l)}_{m,k} = 
 \begin{pmatrix}
  a_{1,1} & a_{1,2} & \cdots & a_{1,k} \\
  a_{2,1} & a_{2,2} & \cdots & a_{2,k} \\
  \vdots  & \vdots  & \ddots & \vdots  \\
  a_{m,1} & a_{m,2} & \cdots & a_{m,k} 
 \end{pmatrix}
\end{equation}
where $M^{(l)}_{i,:k}=\left[|Y^{(l)}_i-Y^{(l)}_1|, |Y^{(l)}_i-Y^{(l)}_2|, \cdots, |Y^{(l)}_i-Y^{(l)}_N|\right]_{:k}$ such that $a_{i,j}\leq a_{i,j+1}$, where $k\leq N$ and $[\cdots]_{:k}$ denotes array slicing up to the $k$th element. Intuitively, each row is a vector whose elements are the magnitudes of the differences a specific datapoint has with its closest-$k$ neighbors in its corresponding distribution. 

This information serves as the feature vectors of the datapoints for adversarial training. The learning objective then is: $f(\cdot;\theta^f)$ is to generate binding strengths that are closer to their ground truths and also have a similar neighborhood structure as the labels distribution. The resulting composite objective for the minimization operation then takes the form:
\begin{equation}
    L_G = L_G^{MSE} + \lambda L_G^{AL}
\end{equation}
where the adversarial training elements for minimization are:
\begin{align}
    \label{eq:d_loss}
    L_D^{AL} &= \mathbb{E}_{\mathbf{x}\sim p} \left[-log D(\mathbf{x}) \right] + \mathbb{E}_{\mathbf{x}\sim G}\left[-log(1-D(\mathbf{x}))\right] \\
    \label{eq:g_loss}
    L_G^{AL} &= \mathbb{E}_{\mathbf{x}\sim G} \left[-log D(\mathbf{x})\right]
\end{align}
The distributions $p$ and $G$ of equations~\ref{eq:d_loss}-\ref{eq:g_loss} are represented by the matrices constructed from the labels and predicted values, respectively. $\lambda$ is a hyperparameter that is used to control the combination of MSE and AL losses of the generator.

In what follows, we demonstrate how this composite loss, together with the integration of the predefined descriptors and end-to-end descriptor learning, improves the model (generator) skill at predicting the binding strengths between a given compound-target pair.

\section{Experiments Protocol}\label{sec:exp}
In this section, we present the design of our experiments and baselines. We also provide our source code and ancillary files at \url{https://github.com/bbrighttaer/ivpgan}.

\subsection{Datasets and Implementations}\label{sec:data_impl}
The benchmark datasets used are the Metz \cite{Metz2011}, KIBA \cite{Tang2014}, and Davis \cite{Davis2011} datasets as provided by the work in \cite{Feng2018}. In their work, they applied a filter threshold to each dataset for which compounds and targets with total number of samples not above the threshold are removed. 
The summary of these datasets are presented in table~\ref{tab:datasets}.

\begin{table}[]
\caption{Dataset sizes}
\label{tab:datasets}
\centering
\begin{tabular}{|l|l|l|l|l|}
\hline
\textbf{Dataset} & \textbf{\begin{tabular}[c]{@{}l@{}}Number of \\ compounds\\/drugs\end{tabular}} & \textbf{\begin{tabular}[c]{@{}l@{}}Number \\ of targets\end{tabular}} & \textbf{\begin{tabular}[c]{@{}l@{}}Total number\\ of pair samples\end{tabular}} & \textbf{\begin{tabular}[c]{@{}l@{}}Filter \\ threshold \\ used\end{tabular}} \\ \hline
Davis            & 72                                                                            & 442                                                                   & 31824                                                                           & 6                                                                    \\ \hline
Metz             & 1423                                                                          & 170                                                                   & 35259                                                                           & 1                                                                    \\ \hline
KIBA             & 3807                                                                          & 408                                                                   & 160296                                                                          & 6                                                                    \\ \hline
\end{tabular}
\end{table}

In our experiments, the ECFP8~\cite{Rogers2010} and Molecular Graph Convolution (GraphConv)~\cite{Altae-Tran2017} serve as representatives of predefined molecular descriptors and differentiable molecular descriptors, respectively. 
We used the data loading and metrics procedures provided by~\cite{Feng2018} (with modifications where necessary) and implemented all models using the Pytorch framework. All our experiments were spread over the servers described in table~\ref{tab:hdw_specs}. 
\begin{table}[]
\caption{Simulation hardware specifications}
\label{tab:hdw_specs}
\begin{tabular}{|l|l|l|l|l|}
\hline
\multicolumn{1}{|c|}{\textbf{Model}}                               & \multicolumn{1}{c|}{\textbf{\# Cores}} & \multicolumn{1}{c|}{\textbf{\begin{tabular}[c]{@{}c@{}}RAM \\ (GB)\end{tabular}}} & \multicolumn{1}{c|}{\textbf{\begin{tabular}[c]{@{}c@{}}Avail. \\ GPUs\end{tabular}}} & \multicolumn{1}{c|}{\textbf{\begin{tabular}[c]{@{}c@{}}\# GPUs \\ used\end{tabular}}} \\ \hline
\begin{tabular}[c]{@{}l@{}}Intel Xeon \\ CPU E5-2687W\end{tabular} & 48                                     & 128                                                                             & \begin{tabular}[c]{@{}l@{}}1 GeForce\\  GTX 1080\end{tabular}                        & 1                                                                                     \\ \hline
\begin{tabular}[c]{@{}l@{}}Intel Xeon \\ CPU E5-2687W\end{tabular} & 24                                     & 128                                                                             & \begin{tabular}[c]{@{}l@{}}4 GeForce \\ GTX 1080Ti\end{tabular}                      & 2                                                                                     \\ \hline
\end{tabular}
\end{table}

\subsection{Baselines}
We compare our proposal to the parametric models proposed in~\cite{Feng2018}. 
In our implementation of the ECFP-PSC architecture, we used the ECFP8 variant, other than the ECFP4 used in~\cite{Feng2018} and in several previous studies. In our preliminary experiments of comparing ECPF4-PSC and ECFP8-PSC architectures, ECFP8 variant mostly outperformed the ECFP4 model. This finding corroborates the view in~\cite{Rogers2010} that while fewer iterations could lead to good performances in similarity and clustering tasks, activity prediction models tend to perform better with greater structure details.

\subsection{Model Training and Evaluation}\label{sec:train_eval}
In our experiments, we used a 5-fold double Cross Validation (CV) model training approach in which three main splitting schemes were used:
\begin{itemize}
    \item \textbf{Warm split}: Every drug or target in the validation set is encountered in the training set.
    \item \textbf{Cold-drug split}: Compounds in the validation set are absent from the training set.
    \item \textbf{Cold-target split}: Targets in the validation set are absent from the training set.
\end{itemize}
Since \emph{cold-start} predictions are typically found in DTI use cases, the cold splits offer an interesting and more challenging validation schemes for the trained models.

As regards evaluation metrics, we measure the Root Mean Squared Error (RMSE) and Pearson correlation coefficient ($ R^2 $) on the validation sets in each CV-fold. Additionally, we measure the Concordance Index (CI) on the validation set as proposed by \cite{Pahikkala2015a}. 

We follow the averaging CV approach where the reported metrics are the averages across the different folds. We also repeat the CV evaluation for different random seeds to minimize randomness. Consequently, all statistics are also averaged across such seeds. Hyperparameters were searched for on the warm split of the davis dataset using the Bayesian optimization API of scikit-optimize. 

\section{Results and Discussion}\label{sec:res_discu}
In tables~\ref{tab:rmse_results}-\ref{tab:r2_results}, we present the RMSE, CI, and R2 values as measured on the best trained models on each of the benchmark datasets, respectively. The standard deviation is placed beneath the mean value of each case. 

Firstly, model complexity and data sizes were influential in model performances. While the IVPGAN and GraphConv-PSC models took longer training times due to the large number of parameters, ECFP8-PSC took much less time to train. The simplicity of ECFP8-PSC also ensured less overfitting with adequate regularization in the face of fewer samples in a number of cases.

Notwithstanding the foregoing observation, IVPGAN mostly achieved the best results than the baseline models, with the GraphConv-PSC being outperformed by the ECFP8-PSC model. Also, just as there exists a general trend of increasing difficulty of prediction from warm split to cold target split in~\cite{Feng2018}, our implementations experienced this behavior as well albeit, with significant improvements in many cases.

Furthermore, there exist a similar trend across tables~\ref{tab:rmse_results}-\ref{tab:r2_results}. Feng et. al~\cite{Feng2018} observed that, for datasets where there are more samples of compounds than targets, cold drug split performances had the tendency to perform better than cold target splits with the reverse being true. However, in our experiments, we observed that such a relationship may be tenuous, at best, and that model skill seem to depend more on model capacity and hyperparameters used for training. In the case of warm splits results, our experiments align with the observation in~\cite{Feng2018} that they are always better than both cold drug and cold target split performances due to the variation in sample sizes. Indeed, it can be observed that cold target split proved to be the most challenging splitting scheme for all models in our experiments. 

Additionally, on the KIBA dataset, although the number of compounds significantly outweighs that of targets, the difference in cold target and cold drug performances, as measured on IVPGAN, is not as pronounced as seen in the baseline models. Thus, with IVPGAN, diversity in compound features seem more necessary for richer representation of compound-target samples.

While IVPGAN mostly outperformed the baseline models, and especially so in the cold split schemes, it was mostly outperformed on the Metz dataset. Aside a possible overfitting due to sample size (also for GraphConv), the hyperparameters used may be less suited for the Metz dataset and its CV split schemes. 

\begin{table}[]
\caption{Performance of regression on benchmark datasets measured in RMSE.}
\label{tab:rmse_results}
\begin{tabular}{|l|l|l|l|l|}
    \hline
    \multicolumn{5}{|c|}{\textbf{RMSE}}                                                                                                                                                                                                                                                       \\ \hline
    \textbf{Datatset}      & \textbf{CV split type} & \textbf{ECFP8}                                                & \textbf{GraphConv}                                            & \textbf{IVPGAN}                                            \\ \hline
    \multirow{3}{*}{Davis} & Warm                                                             & \begin{tabular}[c]{@{}l@{}}0.2216 \\ $\pm 0.082$\end{tabular} & \begin{tabular}[c]{@{}l@{}}0.3537 \\ $\pm 0.053$\end{tabular} & \begin{tabular}[c]{@{}l@{}}\textbf{0.2014} \\ $\pm 0.043$\end{tabular} \\ \cline{2-5} 
                           & Cold drug                                                        & \begin{tabular}[c]{@{}l@{}}0.3978 \\ $\pm 0.105$\end{tabular} & \begin{tabular}[c]{@{}l@{}}0.4751 \\ $\pm 0.123$\end{tabular} & \begin{tabular}[c]{@{}l@{}}\textbf{0.2895} \\ $\pm 0.163$\end{tabular} \\ \cline{2-5} 
                           & Cold target                                                      & \begin{tabular}[c]{@{}l@{}}0.5517 \\ $\pm 0.088$\end{tabular} & \begin{tabular}[c]{@{}l@{}}0.5752 \\ $\pm 0.101$\end{tabular} & \begin{tabular}[c]{@{}l@{}}\textbf{0.2202} \\ $\pm 0.139$\end{tabular} \\ \hline
    \multirow{3}{*}{Metz}  & Warm                                                             & \begin{tabular}[c]{@{}l@{}}\textbf{0.3321} \\ $\pm 0.057$\end{tabular} & \begin{tabular}[c]{@{}l@{}}0.5537 \\ $\pm 0.033$\end{tabular} & \begin{tabular}[c]{@{}l@{}}0.5529 \\ $\pm 0.033$\end{tabular} \\ \cline{2-5} 
                           & Cold drug                                                        & \begin{tabular}[c]{@{}l@{}}\textbf{0.3778} \\ $\pm 0.097$\end{tabular} & \begin{tabular}[c]{@{}l@{}}0.5711 \\ $\pm 0.057$\end{tabular} & \begin{tabular}[c]{@{}l@{}}0.5477 \\ $\pm 0.064$\end{tabular} \\ \cline{2-5} 
                           & Cold target                                                      & \begin{tabular}[c]{@{}l@{}}0.6998 \\ $\pm 0.065$\end{tabular} & \begin{tabular}[c]{@{}l@{}}0.7398 \\ $\pm 0.047$\end{tabular} & \begin{tabular}[c]{@{}l@{}}\textbf{0.5745} \\ $\pm 0.054$\end{tabular} \\ \hline
    \multirow{3}{*}{KIBA}  & Warm                                                             & \begin{tabular}[c]{@{}l@{}}0.4350 \\ $\pm 0.086$\end{tabular} & \begin{tabular}[c]{@{}l@{}}0.5604 \\ $\pm 0.120$\end{tabular} & \begin{tabular}[c]{@{}l@{}}\textbf{0.4003} \\ $\pm 0.082$\end{tabular} \\ \cline{2-5} 
                           & Cold drug                                                        & \begin{tabular}[c]{@{}l@{}}\textbf{0.4502} \\ $\pm 0.128$\end{tabular} & \begin{tabular}[c]{@{}l@{}}0.552 \\ $\pm 0.156$\end{tabular}  & \begin{tabular}[c]{@{}l@{}}0.4690 \\ $\pm 0.132$\end{tabular} \\ \cline{2-5} 
                           & Cold target                                                      & \begin{tabular}[c]{@{}l@{}}0.6645 \\ $\pm 0.137$\end{tabular} & \begin{tabular}[c]{@{}l@{}}0.7555 \\ $\pm 0.153$\end{tabular} & \begin{tabular}[c]{@{}l@{}}\textbf{0.4486} \\ $\pm 0.106$\end{tabular} \\ \hline
    \end{tabular}
\end{table}

\begin{table}[]
\caption{Performance of regression on benchmark datasets measured in CI}
\label{tab:ci_results}
\begin{tabular}{|l|l|l|l|l|}
    \hline
    \multicolumn{5}{|c|}{\textbf{Concordance Index}}                                                                                                                                                                                                \\ \hline
    \textbf{Dataset}       & \textbf{CV split type} & \textbf{ECFP8}                                                & \textbf{GraphConv}                                            & \textbf{IVPGAN}                                            \\ \hline
    \multirow{3}{*}{Davis} & Warm                   & \begin{tabular}[c]{@{}l@{}}0.9647 \\ $\pm 0.020$\end{tabular} & \begin{tabular}[c]{@{}l@{}}0.9335 \\ $\pm 0.011$\end{tabular} & \begin{tabular}[c]{@{}l@{}}\textbf{0.9729} \\ $\pm 0.008$\end{tabular} \\ \cline{2-5} 
                           & Cold drug              & \begin{tabular}[c]{@{}l@{}}0.9099\\ $\pm 0.049$\end{tabular}  & \begin{tabular}[c]{@{}l@{}}0.8784 \\ $\pm 0.052$\end{tabular} & \begin{tabular}[c]{@{}l@{}}\textbf{0.9493} \\ $\pm 0.044$\end{tabular} \\ \cline{2-5} 
                           & Cold target            & \begin{tabular}[c]{@{}l@{}}0.8683 \\ $\pm 0.033$\end{tabular} & \begin{tabular}[c]{@{}l@{}}0.8480 \\ $\pm 0.038$\end{tabular} & \begin{tabular}[c]{@{}l@{}}\textbf{0.9631} \\ $\pm 0.036$\end{tabular} \\ \hline
    \multirow{3}{*}{Metz}  & Warm                   & \begin{tabular}[c]{@{}l@{}}\textbf{0.8923} \\ $\pm 0.025$\end{tabular} & \begin{tabular}[c]{@{}l@{}}0.7968 \\ $\pm 0.027$\end{tabular} & \begin{tabular}[c]{@{}l@{}}0.7913 \\ $\pm 0.029$\end{tabular} \\ \cline{2-5} 
                           & Cold drug              & \begin{tabular}[c]{@{}l@{}}\textbf{0.8730}\\ $\pm 0.044$\end{tabular}  & \begin{tabular}[c]{@{}l@{}}0.7850 \\ $\pm 0.040$\end{tabular} & \begin{tabular}[c]{@{}l@{}}0.7894 \\ $\pm 0.042$\end{tabular} \\ \cline{2-5} 
                           & Cold target            & \begin{tabular}[c]{@{}l@{}}0.7304 \\ $\pm 0.039$\end{tabular} & \begin{tabular}[c]{@{}l@{}}0.7084 \\ $\pm 0.041$\end{tabular} & \begin{tabular}[c]{@{}l@{}}\textbf{0.7776} \\ $\pm 0.038$\end{tabular} \\ \hline
    \multirow{3}{*}{KIBA}  & Warm                   & \begin{tabular}[c]{@{}l@{}}0.8322\\ $\pm 0.024$\end{tabular}  & \begin{tabular}[c]{@{}l@{}}0.7873 \\ $\pm 0.029$\end{tabular} & \begin{tabular}[c]{@{}l@{}}\textbf{0.8433} \\ $\pm 0.023$\end{tabular} \\ \cline{2-5} 
                           & Cold drug              & \begin{tabular}[c]{@{}l@{}}\textbf{0.8132} \\ $\pm 0.047$\end{tabular} & \begin{tabular}[c]{@{}l@{}}0.7736 \\ $\pm 0.048$\end{tabular} & \begin{tabular}[c]{@{}l@{}}0.8070 \\ $\pm 0.051$\end{tabular} \\ \cline{2-5} 
                           & Cold target            & \begin{tabular}[c]{@{}l@{}}0.7185\\ $\pm 0.044$\end{tabular}  & \begin{tabular}[c]{@{}l@{}}0.6661 \\ $\pm 0.052$\end{tabular} & \begin{tabular}[c]{@{}l@{}}\textbf{0.8234} \\ $\pm 0.044$\end{tabular} \\ \hline
    \end{tabular}
\end{table}

\begin{table}[]
\caption{Performance of regression on benchmark datasets measured in $R^2$.}
\label{tab:r2_results}
\begin{tabular}{|l|l|l|l|l|}
    \hline
    \multicolumn{5}{|c|}{\textbf{R2}}                                                                                                                                                                                                                                                         \\ \hline
    \textbf{Dataset}       & \textbf{CV split type} & \textbf{ECFP8}                                                & \textbf{GraphConv}                                            & \textbf{IVPGAN}                                            \\ \hline
    \multirow{3}{*}{Davis} & Warm                                                             & \begin{tabular}[c]{@{}l@{}}0.9252 \\ $\pm 0.061$\end{tabular} & \begin{tabular}[c]{@{}l@{}}0.8254 \\ $\pm 0.039$\end{tabular} & \begin{tabular}[c]{@{}l@{}}\textbf{0.9449} \\ $\pm 0.021$\end{tabular} \\ \cline{2-5} 
                           & Cold drug                                                        & \begin{tabular}[c]{@{}l@{}}0.7573 \\ $\pm 0.171$\end{tabular} & \begin{tabular}[c]{@{}l@{}}0.6773 \\ $\pm 0.159$\end{tabular} & \begin{tabular}[c]{@{}l@{}}\textbf{0.8635} \\ $\pm 0.151$\end{tabular} \\ \cline{2-5} 
                           & Cold target                                                      & \begin{tabular}[c]{@{}l@{}}0.5916 \\ $\pm 0.120$\end{tabular} & \begin{tabular}[c]{@{}l@{}}0.5423 \\ $\pm 0.121$\end{tabular} & \begin{tabular}[c]{@{}l@{}}\textbf{0.9059} \\ $\pm 0.121$\end{tabular} \\ \hline
    \multirow{3}{*}{Metz}  & Warm                                                             & \begin{tabular}[c]{@{}l@{}}\textbf{0.8637} \\ $\pm 0.057$\end{tabular} & \begin{tabular}[c]{@{}l@{}}0.6279 \\ $\pm 0.075$\end{tabular} & \begin{tabular}[c]{@{}l@{}}0.6285 \\ $\pm 0.078$\end{tabular} \\ \cline{2-5} 
                           & Cold drug                                                        & \begin{tabular}[c]{@{}l@{}}\textbf{0.8124} \\ $\pm 0.117$\end{tabular} & \begin{tabular}[c]{@{}l@{}}0.5860 \\ $\pm 0.120$\end{tabular} & \begin{tabular}[c]{@{}l@{}}0.6166 \\ $\pm 0.120$\end{tabular} \\ \cline{2-5} 
                           & Cold target                                                      & \begin{tabular}[c]{@{}l@{}}0.4259 \\ $\pm 0.121$\end{tabular} & \begin{tabular}[c]{@{}l@{}}0.3619 \\ $\pm 0.112$\end{tabular} & \begin{tabular}[c]{@{}l@{}}\textbf{0.5931} \\ $\pm 0.106$\end{tabular} \\ \hline
    \multirow{3}{*}{KIBA}  & Warm                                                             & \begin{tabular}[c]{@{}l@{}}0.7212 \\ $\pm 0.072$\end{tabular} & \begin{tabular}[c]{@{}l@{}}0.5513 \\ $\pm 0.097$\end{tabular} & \begin{tabular}[c]{@{}l@{}}\textbf{0.7658} \\ $\pm 0.065$\end{tabular} \\ \cline{2-5} 
                           & Cold drug                                                        & \begin{tabular}[c]{@{}l@{}}\textbf{0.6677} \\ $\pm 0.137$\end{tabular} & \begin{tabular}[c]{@{}l@{}}0.5026 \\ $\pm 0.152$\end{tabular} & \begin{tabular}[c]{@{}l@{}}0.6475 \\ $\pm 0.142$\end{tabular} \\ \cline{2-5} 
                           & Cold target                                                      & \begin{tabular}[c]{@{}l@{}}0.3648 \\ $\pm 0.128$\end{tabular} & \begin{tabular}[c]{@{}l@{}}0.1910 \\ $\pm 0.088$\end{tabular} & \begin{tabular}[c]{@{}l@{}}\textbf{0.7056} \\ $\pm 0.113$\end{tabular} \\ \hline 
    \end{tabular}
\end{table}

We also conducted qualitative investigations into the prediction performances of all models trained herein. The resulting scatter and joint plots could be seen at \url{https://github.com/bbrighttaer/ivpgan}.
These plots align with the trends identified in tables~\ref{tab:rmse_results}-\ref{tab:r2_results}. We realized that IVPGAN is able to learn the labels distribution better than the baseline models in most of the cases and more so in the cold target CV schemes when there are significantly fewer targets. The realization that the GraphConv-PSC model is not able to properly model the target distribution for smaller datasets, with the ECFP8-PSC model being mostly less so, highlights the potential of IVPGAN. 

The neighborhood alignment dataset in the adversarial training also enables the reduction of residues in most of the scatter plots. 
In summary, these results and findings demonstrate the feasibility and effectiveness of our approach to DTI prediction.

\section{Conclusion}\label{sec:conc}
In this study, we have discussed the use of DL models in DTI prediction and emphasized the significance of the choice of featurization schemes in model training. Using the ECFP8-PSC and GraphConv-PSC models as baselines, we have demonstrated that the IVPGAN approach results in improved model skill in most challenging use cases such as cold target splits. 

Future studies could examine how coordinated multi-view representation learning mechanisms compare to the joint approach adopted in this study. 
In addition, other proposed GAN training techniques could be adopted to address problems associated with GAN training.

\bibliographystyle{IEEEtran}
\bibliography{references}
\end{document}